\documentclass[conference]{IEEEtran}
\usepackage{times}

\usepackage[numbers]{natbib}
\usepackage{multicol}
\usepackage{amssymb}
\usepackage{graphicx}
\usepackage{verbatim}
\usepackage{hyperref}
\usepackage{amsmath}
\usepackage{subcaption}
\usepackage{setspace}
\usepackage{tikz}
\usetikzlibrary{arrows,shapes}

\usepackage{url}  
\newcommand{\youtubeurl}{https://youtu.be/m5z0aFZgEX8}

\begin{document}

% paper title
\title{Detection and Tracking of Liquids \\with Fully Convolutional Networks}

% You will get a Paper-ID when submitting a pdf file to the conference system
\author{Connor Schenck, Dieter Fox\\University of Washington\\\{schenckc,fox\}@cs.washington.edu}

\maketitle

\begin{abstract}
Recent advances in AI and robotics have claimed many incredible results with deep learning, yet no work to date has applied deep learning to the problem of liquid perception and reasoning. In this paper, we apply fully-convolutional deep neural networks to the tasks of detecting and tracking liquids. We evaluate three models: a single-frame network, multi-frame network, and a LSTM recurrent network. Our results show that the best liquid detection results are achieved when aggregating data over multiple frames, in contrast to standard image segmentation. They also show that the LSTM network outperforms the other two in both tasks. This suggests that LSTM-based neural networks have the potential to be a key component for enabling robots to handle liquids using robust, closed-loop controllers.
\end{abstract}

\IEEEpeerreviewmaketitle

\section{Introduction}

To robustly handle liquids, such as pouring a certain amount of water into a
bowl, a robot must be able to perceive and reason about liquids in a way that
allows for closed-loop control. Liquids present many challenges compared to
solid objects. For example, liquids can not be interacted with directly by a
robot, instead the robot must use a tool or container; often containers
containing some amount of liquid are opaque, obstructing the robot's view of the
liquid and forcing it to remember the liquid in the container, rather than
re-perceiving it at each timestep; and finally liquids are frequently
transparent, making simply distinguishing them from the background a difficult
task. Taken together, these challenges make perceiving and manipulating liquids
highly non-trivial.

Recent advances in deep learning have enabled a leap in performance not only on visual
recognition tasks, but also in areas ranging from playing Atari games
\cite{guo2014} to end-to-end policy training in robotics \cite{levine2015}.  In
this paper, we investigate how deep learning techniques can be used for
perceiving liquids during pouring tasks. We develop a method for generating
large amounts of labeled pouring data for training and testing using a realistic
liquid simulation and rendering engine, which we use to generate a data set with
over 4.5 million labeled images. Using this dataset, we evaluate multiple deep
learning network architectures on the tasks of detecting liquid in an image and
tracking the location of liquid even when occluded.

The rest of this paper is laid out as follows. Section \ref{sec:rel_work} describes work related to ours. Section \ref{sec:methodology} details our experimental methodology. Section \ref{sec:evaluation} describes how we evaluated the neural networks. Section \ref{sec:results} details our results. Section \ref{sec:conclusion} contains a discussion of the implications of the results and the conclusions that can be drawn from them. And finally section \ref{sec:future_work} details our directions for future work.

\section{Related Work}
\label{sec:rel_work}

To the best of our knowledge, no prior work has investigated directly perceiving
and reasoning about liquids. Existing work relating to liquids either uses
coarse simulations that are disconnected to real liquid perception and dynamics
\cite{kunze2015,yamaguchi2015} or constrained task spaces that bypass
the need to perceive or reason directly about liquids
\cite{langsfeld2014,okada2006,tamosiunaite2011,cakmak2012,rozo2013}. While some
of this work has dealt with pouring, none of it has attempted to directly
perceive the liquids from raw sensory data, in contrast to this paper, in which
we investigate ways to do just that.

Although there has been some prior work in robotics that has dealt with perception and liquids, though in constrained task spaces.
Work by Rankin {\it et al.} \cite{rankin2010,rankin2011} investigated ways to detect pools of water for an unmanned ground vehicle navigating rough terrain. 
However they detected water based on simple color features or sky reflections, and didn't reason about the dynamics of the water, instead treating it as a static obstacle. 
Griffith {\it et al.} \cite{griffith2012} used the auditory and proprioceptive feedback from objects interacted with in a sink environment with a running water tap in order to learn about those objects, although in this case the robot did not detect or reason about the water, rather it used the water as a means to learn about and categorize other objects. 
In contrast to \cite{griffith2012}, we use vision to directly detect the liquid itself, and unlike \cite{rankin2010,rankin2011}, we treat the liquid as dynamic and reason about it, rather than treating it as a static obstacle.

In order to perceive liquids at the pixel level, we make use of
fully-convolutional neural networks (FCN). FCNs have been successfully applied
to the task of image segmentation in the past
\cite{long2015,havaei2015,romera2015} and are a natural fit for pixel-wise
classification. In addition to FCNs, we utilize long short-term memory (LSTM)
\cite{hochreiter1997} recurrent cells to reason about the temporal evolution of
liquids. LSTMs are preferable over more standard recurrent networks for
long-term memory as they overcome many of the numerical issues during training
such as exploding gradients \cite{greff2015}. LSTM-based CNNs have been
successfully applied to many temporal memory tasks by previous work
\cite{junhyuk2015,romera2015}, and in fact \cite{romera2015} even combine LSTMs
and FCNs by replacing the standard fully-connected layers of their LSTMs with
$1\times1$ convolution layers. We use a similar method in this paper.

\section{Methodology}
\label{sec:methodology}

In order to train neural networks to perceive and reason about liquids, we must first have labeled data to train on. 
Getting pixel-wise labels for real-world data can be difficult, so in this paper we opt to use a realistic liquid simulator. 
In this way we can acquire ground truth pixel labels while generating images that appear as realistic as possible.
We train three different types of convolutional neural networks (CNNs) on this generated data to detect and track the liquid: single-frame CNN, multi-frame CNN, and LSTM-CNN.

\subsection{Data Generation}
\label{sec:data_gen}

\begin{figure}
    \centering
    \setlength{\fboxsep}{0pt}
    \setlength{\fboxrule}{1pt}
    \setlength{\unitlength}{1.0cm}
    \begin{picture}(8.5,8.5)
        \put(0.0,0.0){\fbox{\includegraphics[width=8.5cm]{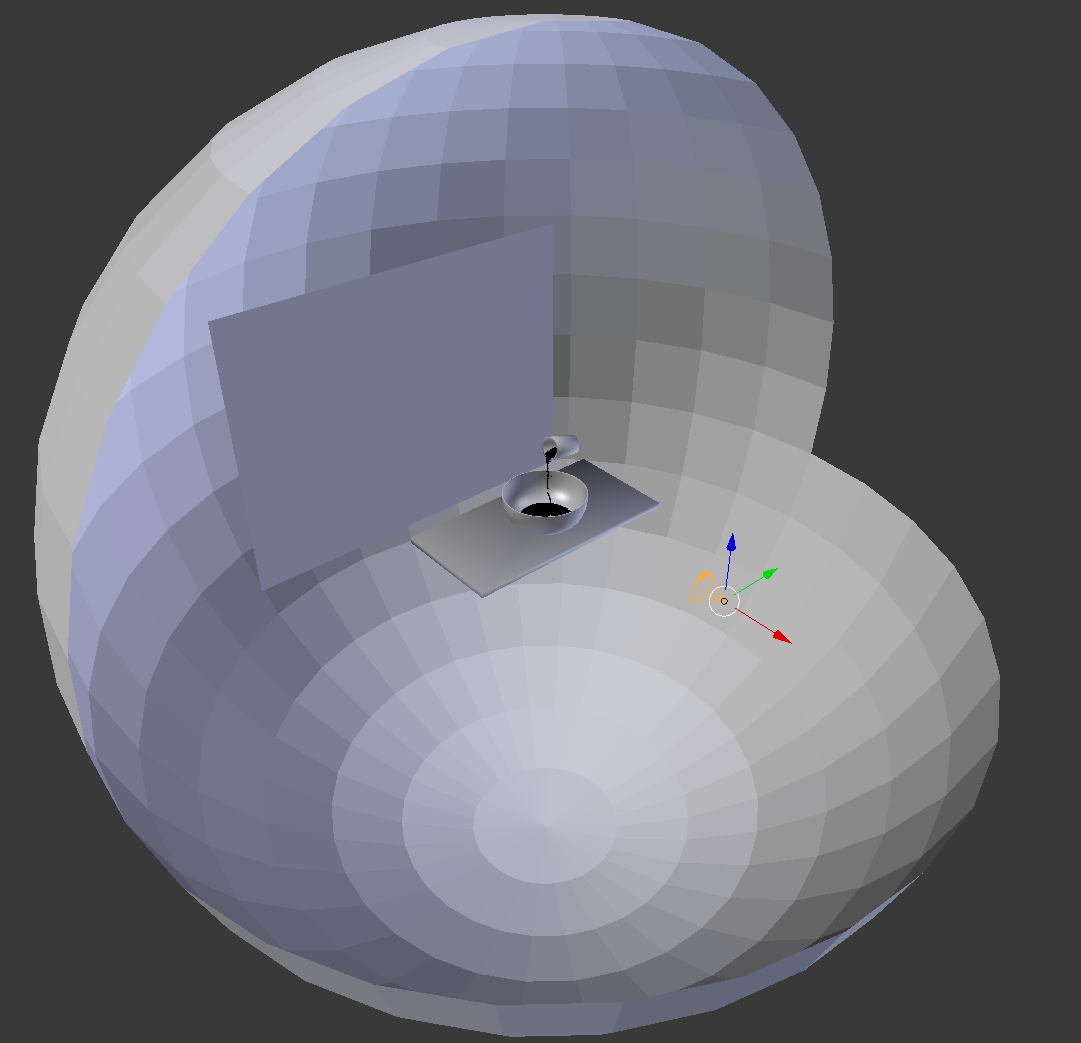}}}
        %\put(0.0,0.0){\drawbox{5.0cm}{3.5cm}}
    \end{picture}
    \caption{The setup used to simulate and render liquid sequences. The objects are shown here textureless for clarity. The sphere surrounding all the objects has been cut away to allow viewing of the objects inside. The orange shape represents the camera's viewpoint, and the flat plane across the table from it is the plane on which the video sequence is rendered. Note that this plane is sized to exactly fill the camera's view frustum. The background sphere is not directly visible by the camera and is used primarily to compute realistic reflections.}
    \label{fig:blender_scene}
\end{figure}

\begin{figure}
    \centering
    \setlength{\fboxsep}{0pt}
    \setlength{\fboxrule}{1pt}
    \begin{subfigure}{7.0cm}
        \setlength{\unitlength}{1.0cm}
        \begin{picture}(7.0,5.5)
            \put(0.0,0.0){\fbox{\includegraphics[width=7.0cm]{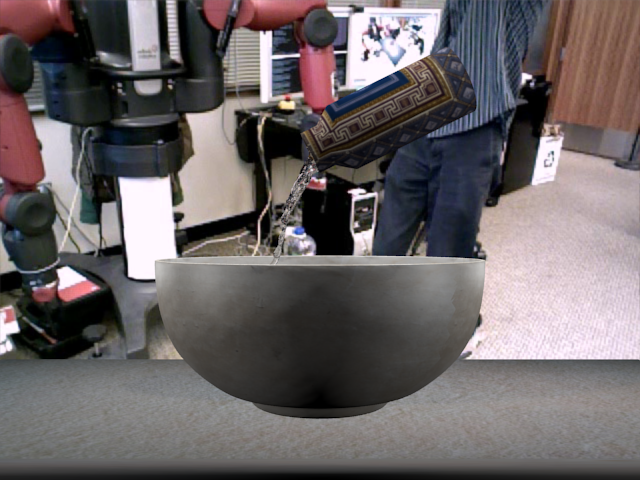}}}
            %\put(0.0,0.0){\drawbox{5.0cm}{3.5cm}}
        \end{picture}
    \end{subfigure}
    
    \begin{subfigure}{7.0cm}
        \setlength{\unitlength}{1.0cm}
        \begin{picture}(7.0,5.5)
            \put(0.0,0.0){\fbox{\includegraphics[width=7.0cm]{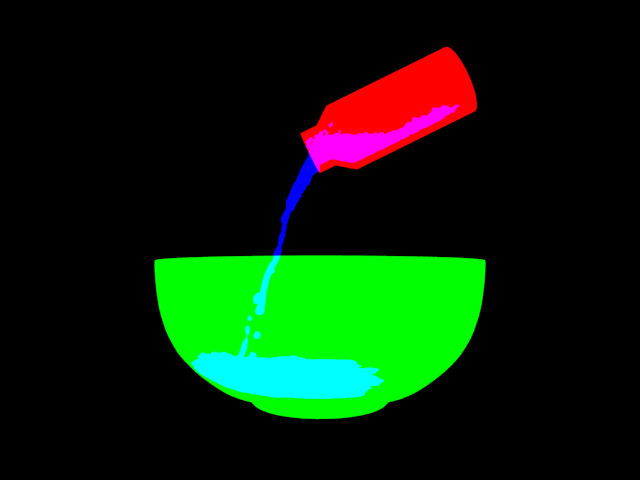}}}
            %\put(0.0,0.0){\drawbox{5.0cm}{3.5cm}}
        \end{picture}
    \end{subfigure}
    \caption{ An example of a frame rendered by our data generation algorithm. The upper image is the raw RGB image generated by the renderer. The lower image is the ground truth binary pixel labels, where the blue channel labels the liquid pixels, the green channel the bowl, and the red channel the cup. The alpha channel (not shown) indicates which of the three (liquid, bowl, cup), if any, is visible at that pixel.}
    \label{fig:data_gen}
\end{figure}

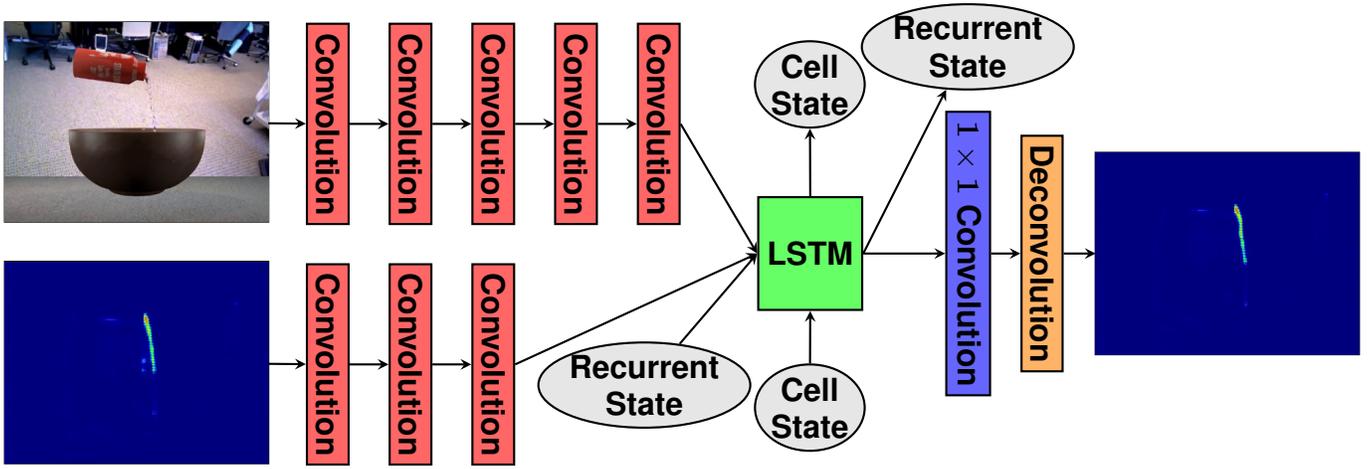
\begin{figure*}[t]
\begin{tikzpicture}[->,>=stealth,auto,node distance=1.6cm,thick,
  input node/.style={rectangle,draw,anchor=west,align=center,inner sep=0,outer sep=0},
  conv node/.style={rectangle,fill=red!60,draw,font=\sffamily\large\bfseries,align=center,anchor=west,minimum height=1.5cm},
  blob node/.style={ellipse,fill=gray!20,draw,font=\sffamily\large\bfseries,align=center,inner sep=0}]

  \node[input node] (In3) at (0.0,5.5) {\includegraphics[width=3.5cm]{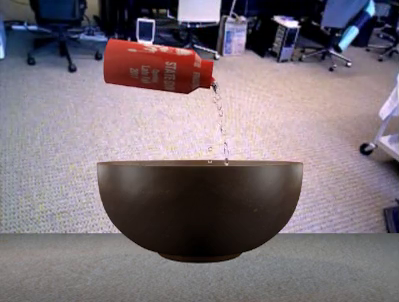}};
  \node[input node] (Rec1) at (0.0,2.3) {\includegraphics[width=3.5cm]{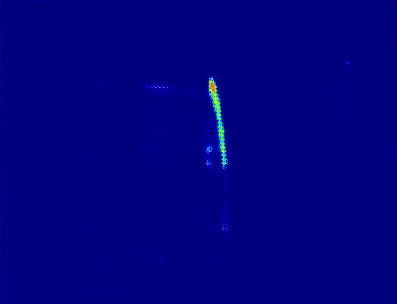}};

  \node[conv node] (Conv1) at (4.0,5.475) {\rotatebox{270}{Convolution}};
  \node[conv node] (Conv2) at (5.1,5.475) {\rotatebox{270}{Convolution}};
  \node[conv node] (Conv3) at (6.2,5.475) {\rotatebox{270}{Convolution}};
  \node[conv node] (Conv4) at (7.3,5.475) {\rotatebox{270}{Convolution}};
  \node[conv node] (Conv5) at (8.4,5.475) {\rotatebox{270}{Convolution}};
  
  \node[conv node] (rec_conv1) at (4.0,2.275) {\rotatebox{270}{Convolution}};
  \node[conv node] (rec_conv2) at (5.1,2.275) {\rotatebox{270}{Convolution}};
  \node[conv node] (rec_conv3) at (6.2,2.275) {\rotatebox{270}{Convolution}};
  
  \node[conv node] (lstm1) at (10.0,3.75) [fill=green!60]{LSTM};
  \node[conv node] (fc_conv1) at (12.5,3.75) [fill=blue!60]{\rotatebox{270}{$\mathbf{1 \times 1}$ Convolution}};
  \node[conv node] (deconv) at (13.5,3.75) [fill=orange!60]{\rotatebox{270}{Deconvolution}};
  \node[input node] (Out1) at (14.5,3.75) {\includegraphics[width=3.5cm]{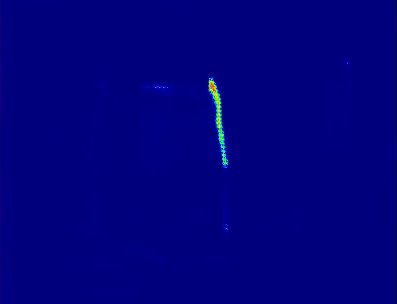}};

  \node[blob node] (rec_in2) at (8.5, 2.0) {Recurrent\\State};
  \node[blob node] (rec_in3) at (10.7, 1.7) {Cell\\State};
  
  \node[blob node] (rec_out2) at (12.8, 6.5) {Recurrent\\State};
  \node[blob node] (rec_out3) at (10.7, 6.0) {Cell\\State};
  
  \draw (In3) -- (Conv1);
  \draw (Conv1) -- (Conv2);
  \draw (Conv2) -- (Conv3);
  \draw (Conv3) -- (Conv4);
  \draw (Conv4) -- (Conv5);
  
  \draw (Rec1) -- (rec_conv1);
  \draw (rec_conv1) -- (rec_conv2);
  \draw (rec_conv2) -- (rec_conv3);

  \draw (Conv5.east) -- (lstm1.west);
  \draw (rec_conv3.east) -- (lstm1.west);
  
  \draw (lstm1) -- (fc_conv1);
  \draw (fc_conv1) -- (deconv);
  \draw (deconv) -- (Out1);
  
  \draw (rec_in2) -- (lstm1.west);

  \draw (rec_in3) -- (lstm1.south);
  \draw (lstm1.east) -- (rec_out2);
  \draw (lstm1.north) -- (rec_out3);
\end{tikzpicture}
\caption{ Layout of the LSTM-CNN. It takes as input the current frame as well as its own predictions from the previous timestep. During training we initialize this at the first timestep as ground truth, but during testing we initialize it as all zeros. The LSTM takes as recurrent input its own output from the previous timestep and the cell state. Refer to figure 1 of \cite{greff2015} for more details. Each of the convolution layers is followed by rectified linear and max pooling layers. The $1 \times 1$ convolution layer is followed by a rectified linear layer.}
\label{fig:network}
\end{figure*}

We generate data using the 3D-modeling application Blender \cite{blender2016} and the library El'Beem for liquid simulation, which is based on the lattice-Boltzmann method for efficient, physically accurate liquid simulations \cite{korner2006}. 
We separate the data generation process into two steps: simulation and rendering. 
During simulation, the liquid simulator calculates the trajectory of the surface mesh of the liquid as the cup pours the liquid into the bowl. 
%We varied the parameters of the simulation (e.g., cup type) to create 81 different simulations, with each simulation lasting exactly 15 seconds for a total of 450 frames.
We vary 4 variables during simulation: the type of cup (cup, bottle, mug), the type of bowl (bowl, dog dish, fruit bowl), the initial amount of liquid (30\% full, 60\% full, 90\% full), and the pouring trajectory (slow, fast, partial), for a total of 81 simulations. Each simulation lasts exactly 15 seconds for a total of 450 frames (30 frames per second).

Next we render each simulation. 
We separate simulation from rendering because it allows us to vary other variables that don't affect the trajectory of the liquid mesh (e.g., camera viewpoint), which provides a significant speedup as liquid simulation is much more computationally intensive than rendering. 
In order to approximate realistic reflections, we mapped a 3D photo sphere image taken in our lab to the inside of a sphere, which we placed in the scene surrounding all the objects. 
%We mapped a 3D photo sphere image taken in our lab to the inside of a sphere, which we placed in the scene surrounding all the objects, and we placed a plane in front of the camera but behind the objects that plays a video of activity in our lab.
To prevent overfitting to a static background, we also add a plane in the image in front of the camera and behind the objects that plays a video of activity in our lab that approximately matches with that location in the background sphere. 
This setup is shown in figure \ref{fig:blender_scene}.
The liquid is always rendered as 100\% transparent, with only reflections, refractions, and specularities differentiating it from the background. 
%We render the liquid as 100\% transparent and varied the render parameters (e.g., camera viewpoint) to randomly generate 10,122 sequences from the 81 simulations for a total of 4,554,900 images.
For each simulation, we vary 6 variables: camera viewpoint (48 preset viewpoints), background video (8 videos), cup and bowl textures (6 textures each), liquid reflectivity (normal, none), and liquid index-of-refraction (air-like, low-water, normal-water). Additionally, we also generate negative examples without liquid. In total, this yields 165,888 possible renders for each simulation. It is infeasible to render them all, so we randomly sample variable values to render.

The labels are generated for each object (liquid, cup, bowl) as follows.
First, all other objects in the scene are set to render as invisible.
Next, the material for the object is set to render as a specific, solid color, ignoring lighting.
The sequence is then rendered, yielding a class label for the object for each pixel.
An example of labeled data and its corresponding rendered image is shown in figure \ref{fig:data_gen}. 
The cup, bowl, and liquid are rendered as red, green and blue respectively.
Note that this method allows each pixel to have multiple labels, e.g., some of the pixels in the cup are labeled as both cup and liquid (magenta in the lower part of figure \ref{fig:data_gen}).
To determine which of the objects, if any, is visible at each pixel, we render the sequence once more with all objects set to render as their respective colors, and we use the alpha channel in the ground truth images to encode the visible class label. 

To evaluate our learning architectures, we generated 10,122 pouring sequences by randomly selecting render variables as described above as well as generating negative sequences (i.e., sequences without any water), for a total of 4,554,900 training images. For simplicity, we only used sequences rendered from 6 of the 48 possible camera poses.

\subsection{Detecting and Tracking liquids}

We test three network layouts for the tasks of detecting and tracking liquids: CNN, MF-CNN, and LSTM-CNN. 
%Figure \ref{fig:network} shows the layout of the LSTM-CNN. 
%In the LSTM layer we use $1 \times 1$ convolutions in place of fully-connected layers. 
%The CNN network is similar to the LSTM-CNN, with the LSTM layer replaced by a $1 \times 1$ convolution layer (followed by a rectified linear layer), and the lower-left half of the diagram (the part of the network that processes the previous timestep's output) removed. 
%The multi-frame CNN (MF-CNN) is similar to the CNN, except it process a series of consecutive frames independently through the first layers of the network, and concatenates them prior to the $1 \times 1$ convolution layers. 
%The MF-CNN then predicts at the timestep of the last frame in its input. 
%For the tracking task, we reduce the number of initial convolution layers on the input from 5 to 3 for each network.

\begin{itemize}
\item {\bf CNN} The first layout is a standard convolutional neural network (CNN) with a fixed number of convolutional layers, each followed by a rectified linear layer and a max pooling layer. In place of fully-connected layers, we use two $1 \times 1$ convolutional layers, each followed by a rectified linear layer. The last layer of the network is a deconvolutional layer. 
\item {\bf MF-CNN} The second layout is a multi-frame CNN. Instead of taking in a single frame, it takes as input multiple consecutive frames and predicts at the last frame. Each frame is convolved independently through the first part of the network, which is composed of a fixed number of convolutional layers, each followed by a rectified linear and max pooling layer. The output for each frame is then concatenated together channel-wise, and then fed to two $1 \times 1$ convolutional layers, each followed by a rectified linear layer, and finally a deconvolutional layer. We fix the number of input frames for this layout to 32 for this paper, i.e., approximately 1 second's worth of data (30 frames per second). 
\item {\bf LSTM-CNN} The third layout is similar to the single frame CNN layout, with the first $1 \times 1$ convolutional layer replaced with a LSTM layer (see figure 1 of \cite{greff2015} for a detailed layout of the LSTM layer). We replace the fully-connected layers of a standard LSTM with $1 \times 1$ convolutional layers. The LSTM takes as recurrent input the cell state from the previous timestep, its output from the previous timestep, and the output of the network from the previous timestep processed through 3 convolutional layers (each followed by a rectified linear and max pooling layer). During training, when unrolling the LSTM CNN, we initialize this last recurrent input with the ground truth at the first timestep, but during testing we initialize it with all zeros.
\end{itemize}

Figure \ref{fig:network} shows the layout of the LSTM-CNN. For the tracking task, we reduce the number of initial convolution layers on the input from 5 to 3 for each network.

We use the Caffe deep learning framework \cite{jia2014} to implement our networks.

\begin{comment}

To detect and track the liquids across a pouring sequence, we train a fully-convolutional neural network. The layout we use is shown in figure \ref{fig:network}. The network takes as input the current RGB image and its own predictions from the previous timestep. Each box in figure \ref{fig:network} represents a series of convolution layers, where each convolution layer is followed by a rectified linear layer and then a max pooling layer. We use an LSTM to handle recurrent state, and we use $1 \times 1$ convolutional layers to process the input to each gate of the LSTM (in contrast to fully-connected layers as in \cite{greff2015}). The final section of our network is a series of $1 \times 1$ convolutional layers, which take the place of fully-connected layers in standard cnns, each of which is followed by a rectified linear layer. Finally, we use a deconvolutional layer to expand the size of the output to be the same as the input size.

We split training into two phases. First, we train the network to detect only the visible liquid (that is, pixels with the liquid class label and are also ``on top'' of all other objects). Next, we train the network to track all liquid in the scene regardless if it is visible or not. In both cases the output of the network is a heatmap of where it believes the liquid is, in the first case only the visible liquid, in the second liquid obstructed by the bowl or cup in addition to the visible liquid. We use the Caffe deep learning framework to implement our networks \cite{jia2014}.
\end{comment}

\section{Evaluation}
\label{sec:evaluation}

\begin{figure*}[t]
    \centering
    \setlength{\fboxsep}{0pt}
    \setlength{\fboxrule}{1pt}
    \setlength{\unitlength}{1.0cm}
    \begin{picture}(15.0,11.75)
        \put(0.0,0.0){\fbox{\includegraphics[width=3.0cm]{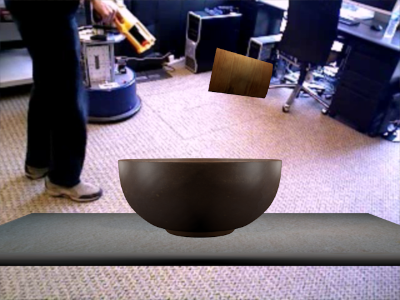}}}
        \put(3.0,0.0){\fbox{\includegraphics[width=3.0cm]{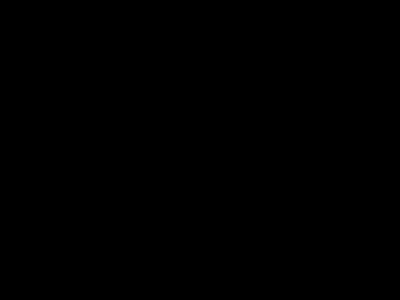}}}
        \put(6.0,0.0){\fbox{\includegraphics[width=3.0cm]{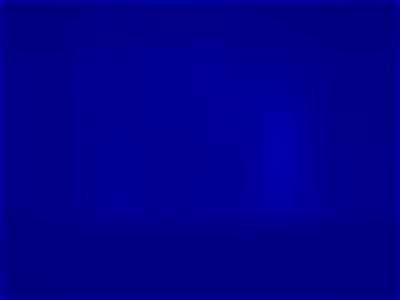}}}
        \put(9.0,0.0){\fbox{\includegraphics[width=3.0cm]{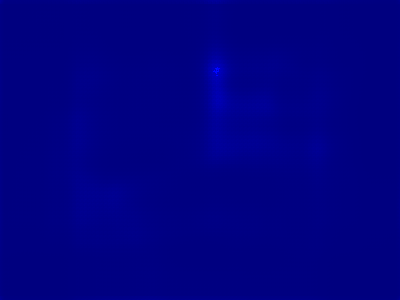}}}
        \put(12.0,0.0){\fbox{\includegraphics[width=3.0cm]{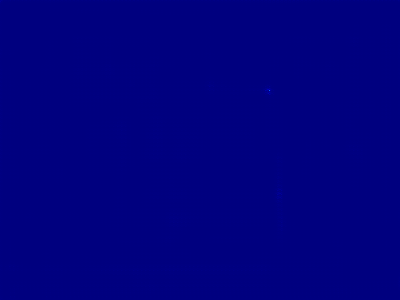}}}
        
        \put(0.0,2.25){\fbox{\includegraphics[width=3.0cm]{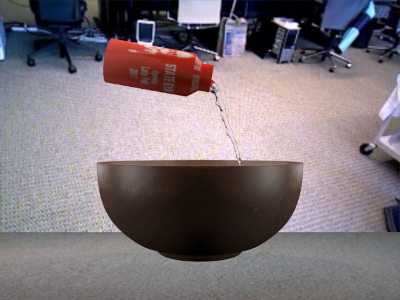}}}
        \put(3.0,2.25){\fbox{\includegraphics[width=3.0cm]{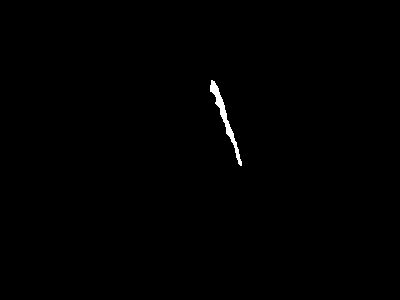}}}
        \put(6.0,2.25){\fbox{\includegraphics[width=3.0cm]{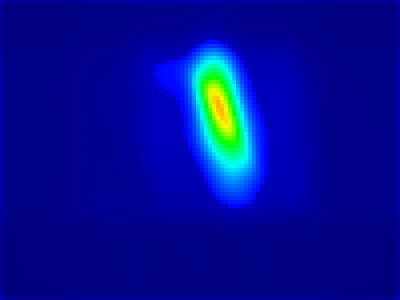}}}
        \put(9.0,2.25){\fbox{\includegraphics[width=3.0cm]{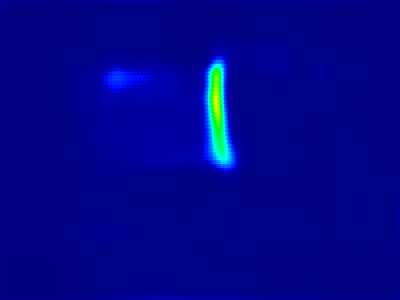}}}
        \put(12.0,2.25){\fbox{\includegraphics[width=3.0cm]{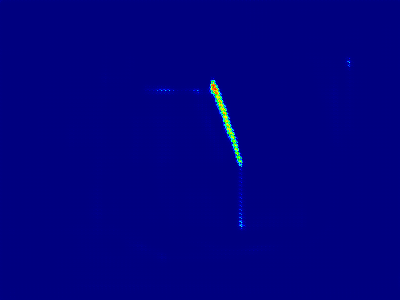}}}

        \put(0.0,4.5){\fbox{\includegraphics[width=3.0cm]{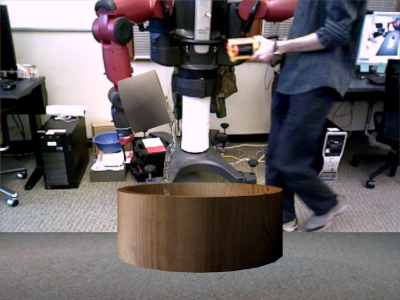}}}
        \put(3.0,4.5){\fbox{\includegraphics[width=3.0cm]{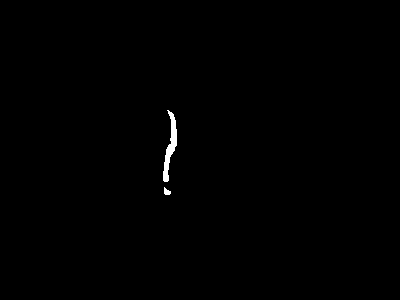}}}
        \put(6.0,4.5){\fbox{\includegraphics[width=3.0cm]{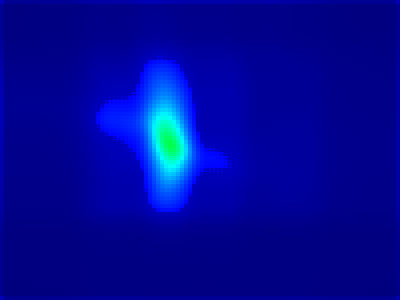}}}
        \put(9.0,4.5){\fbox{\includegraphics[width=3.0cm]{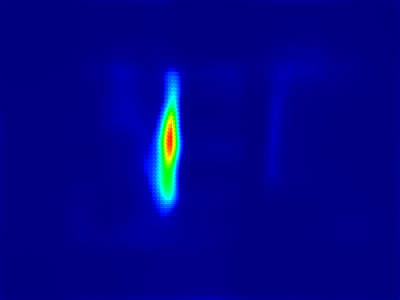}}}
        \put(12.0,4.5){\fbox{\includegraphics[width=3.0cm]{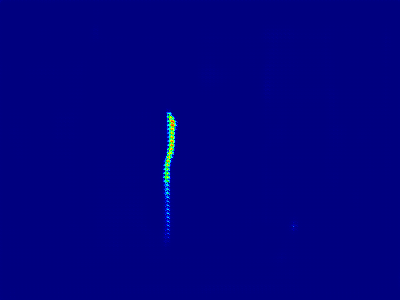}}}
        
        \put(0.0,6.75){\fbox{\includegraphics[width=3.0cm]{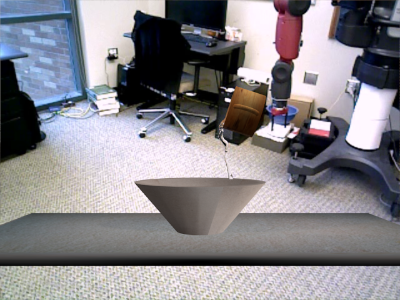}}}
        \put(3.0,6.75){\fbox{\includegraphics[width=3.0cm]{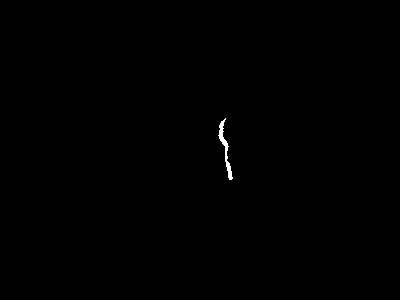}}}
        \put(6.0,6.75){\fbox{\includegraphics[width=3.0cm]{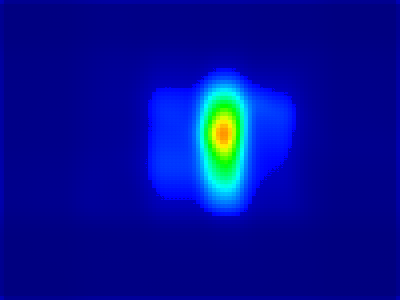}}}
        \put(9.0,6.75){\fbox{\includegraphics[width=3.0cm]{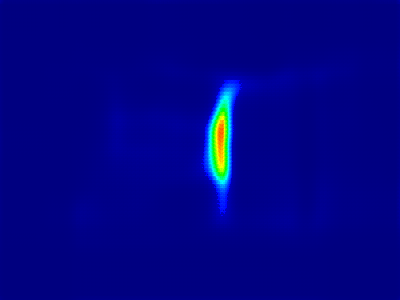}}}
        \put(12.0,6.75){\fbox{\includegraphics[width=3.0cm]{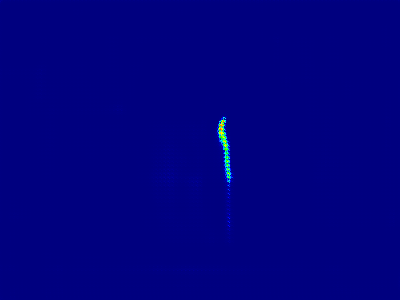}}}
        
        \put(0.0,9.0){\fbox{\includegraphics[width=3.0cm]{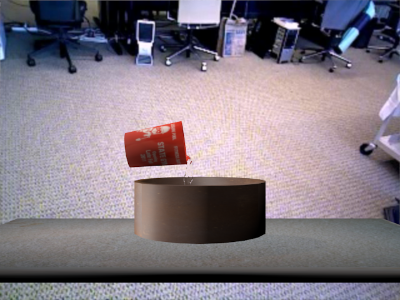}}}
        \put(3.0,9.0){\fbox{\includegraphics[width=3.0cm]{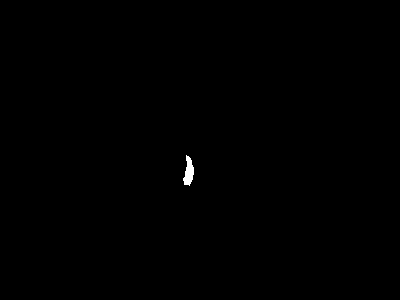}}}
        \put(6.0,9.0){\fbox{\includegraphics[width=3.0cm]{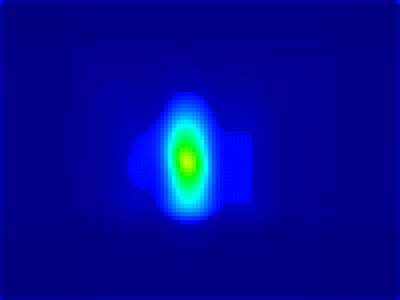}}}
        \put(9.0,9.0){\fbox{\includegraphics[width=3.0cm]{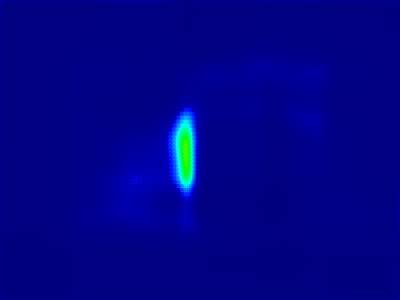}}}
        \put(12.0,9.0){\fbox{\includegraphics[width=3.0cm]{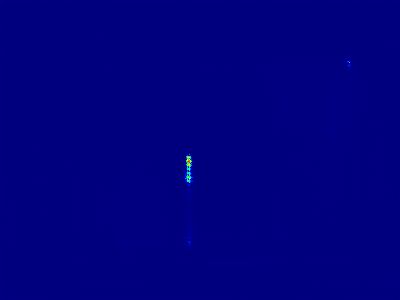}}}
        
        \put(0.7,11.35){{\bf\LARGE Input}}
        \put(3.7,11.35){{\bf\LARGE Labels}}
        \put(7.0,11.35){{\bf\LARGE CNN}}
        \put(9.4,11.35){{\bf\LARGE MF-CNN}}
        \put(12.05,11.35){{\bf\LARGE LSTM-CNN}}
    \end{picture}
    \caption{ Qualitative liquid detection results. The Input column is the input to the networks, the Labels column is the ground truth labeling of each pixel as liquid or not liquid, and the CNN, MF-CNN, and LSTM-CNN columns show a heatmap of the prediction of each network for each of the input frames. 5 sequences were randomly selected from our training set, and the frame with the most liquid pixels was picked for display here, with the exception of the last row, which shows how the networks perform when there is no liquid present.}
    \label{fig:results}
\end{figure*}
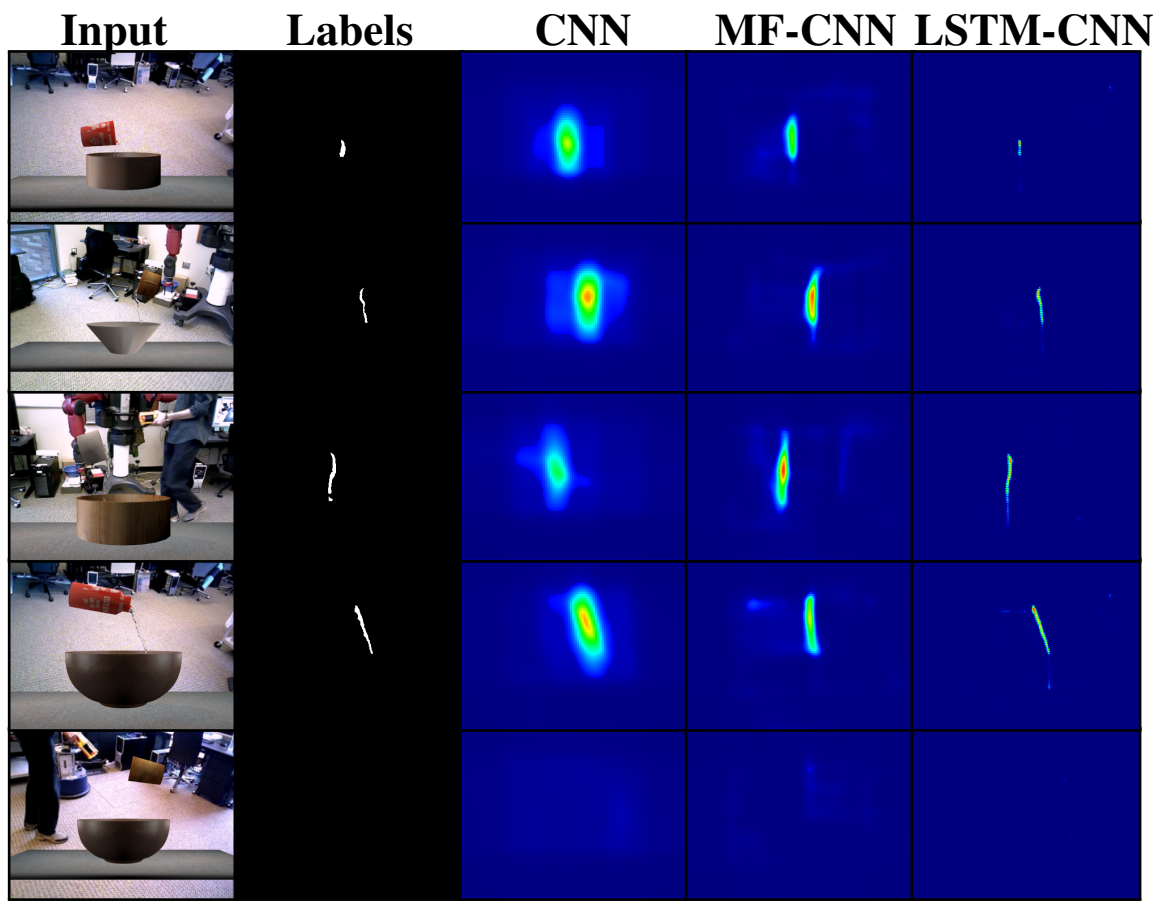

We evaluated the three network types on both the detection and tracking tasks. For detection, the networks were given the full rendered RGB image as input (similar to the top image in figure \ref{fig:data_gen}) at a resolution of $400 \times 300$ pixels. The output was a classification at each pixel as liquid or not liquid. Each network was first trained for 60,000 iterations on image crops of visible liquid, and then again for another 60,000 iterations on the full image (the use of only convolutional layers rather than fully connected layers allows for variable sized inputs and outputs). The weights of the LSTM-CNN were initialized with the weights of the single-frame CNN trained on only cropped image patches. During training, the LSTM-CNN was unrolled for 32 timesteps.

For tracking, the networks were given pre-segmented input images, with the goal being to track the liquid when it is not visible. The input was similar to the bottom image from figure \ref{fig:data_gen}, except that only visible liquid was shown (in the case of figure \ref{fig:data_gen}, the cyan and magenta liquid would not have been shown because it was occluded by the bowl and cup respectively). Because these input images are more structured, we lowered the resolution to $130 \times 100$. The output was the pixel-wise classification of liquid or not liquid, including pixels where the liquid was occluded by other objects in the scene. During training, the LSTM-CNN was unrolled for 180 timesteps.

\section{Results}
\label{sec:results}

\subsection{Detection Results}

\begin{figure}
    \centering
    \setlength{\unitlength}{1.0cm}
    \begin{subfigure}{6.0cm}
        \begin{picture}(6.0,5.0)
            \put(0.0,0.0){\includegraphics[width=6.0cm]{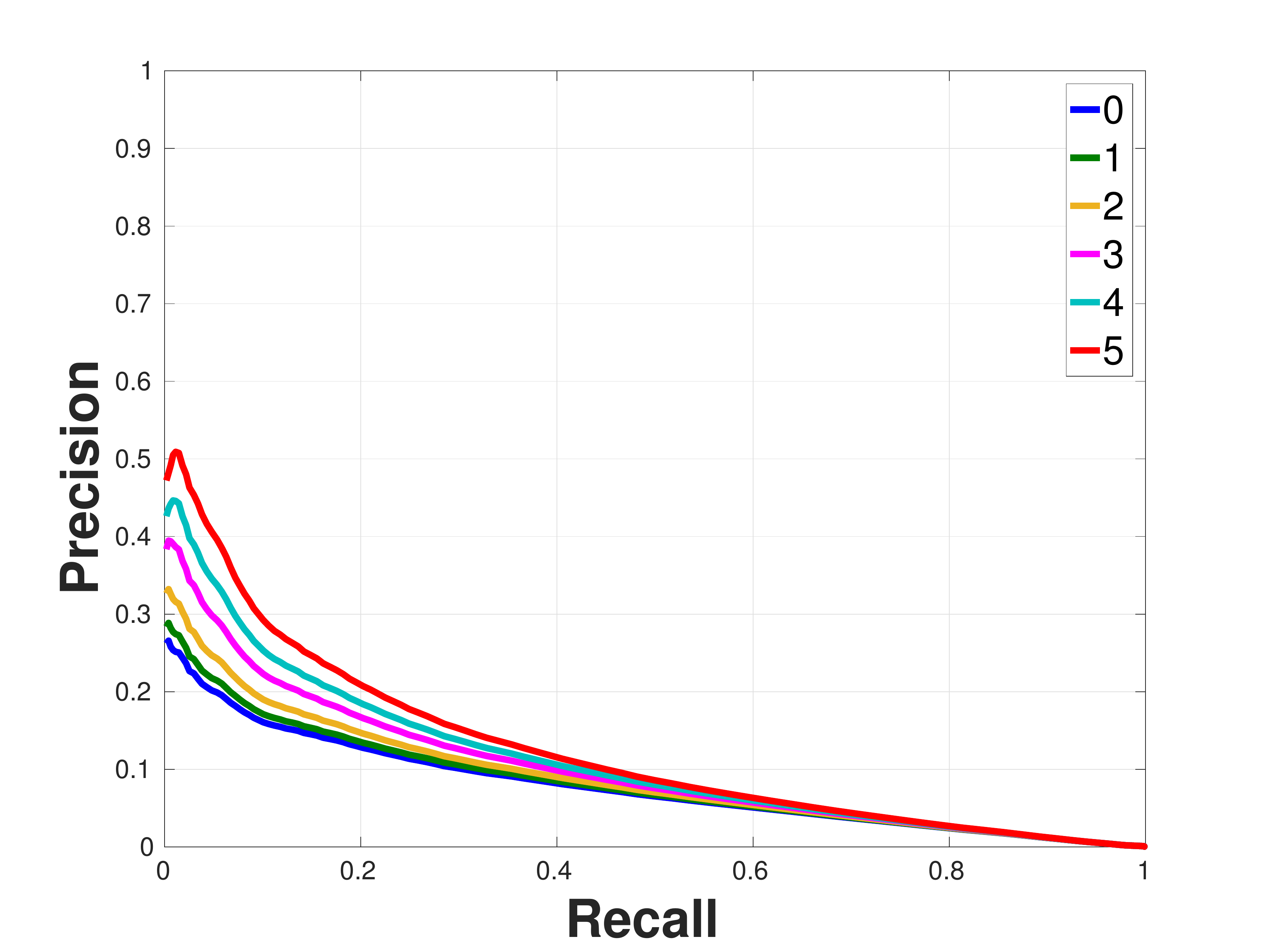}}
        \end{picture}
        \caption{CNN}
        \label{fig:results_detection_cnn}
    \end{subfigure}
    
    \begin{subfigure}{6.0cm}
        \begin{picture}(6.0,5.0)
            \put(0.0,0.0){\includegraphics[width=6.0cm]{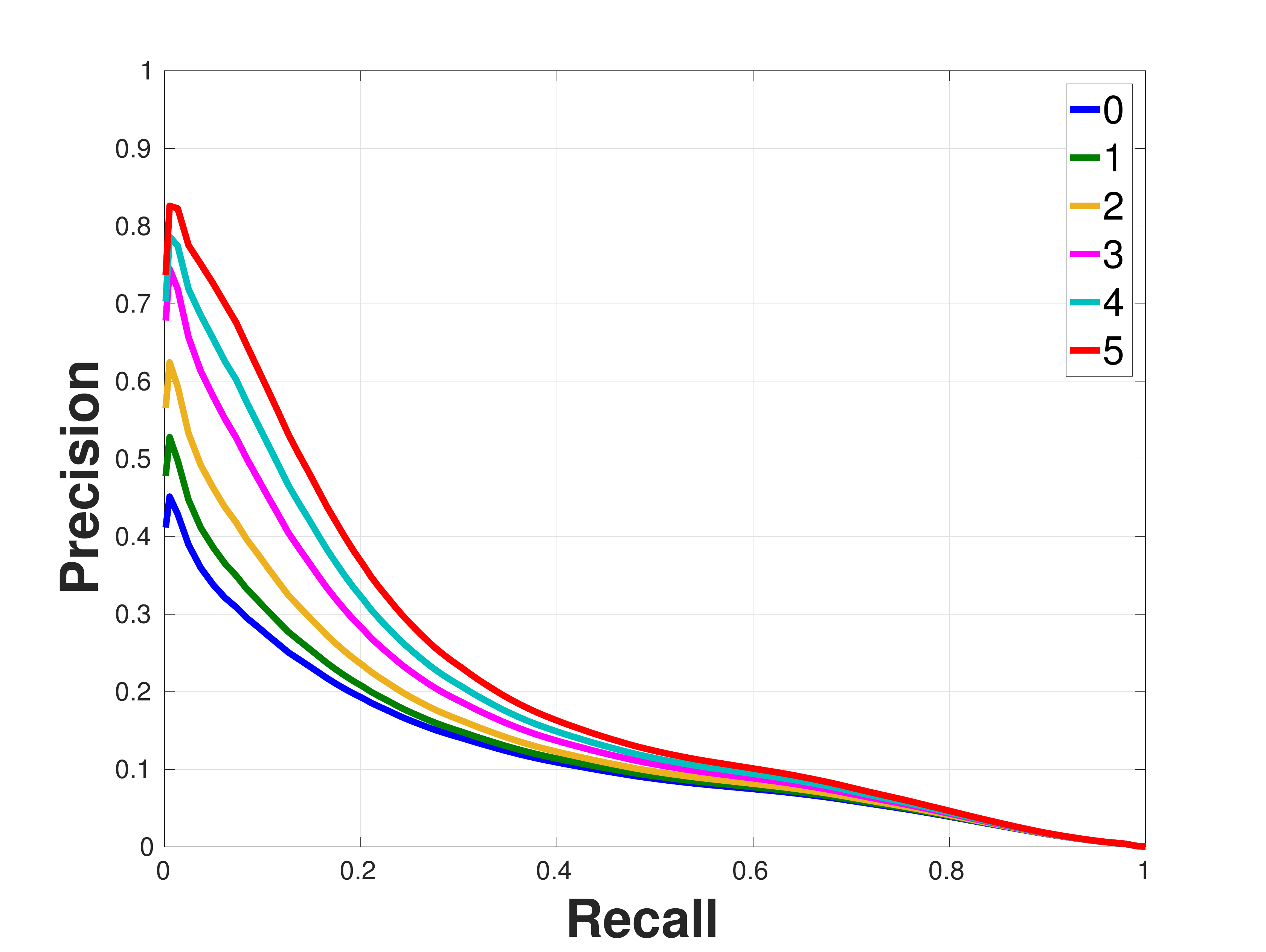}}
        \end{picture}
        \caption{MF-CNN}
        \label{fig:results_detection_wcnn}
    \end{subfigure}
    
    \begin{subfigure}{6.0cm}
        \begin{picture}(6.0,5.0)
            \put(0.0,0.0){\includegraphics[width=6.0cm]{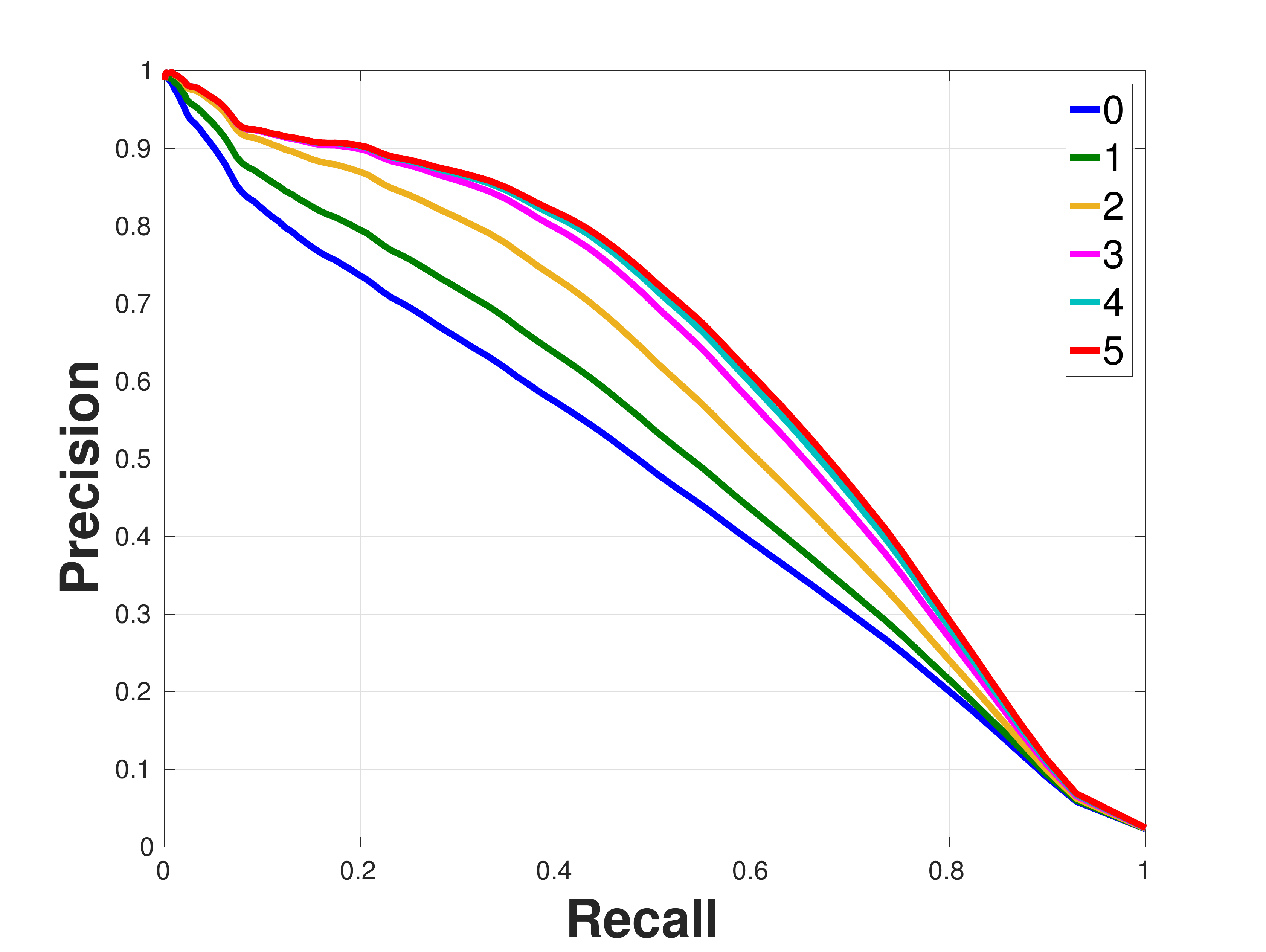}}
        \end{picture}
        \caption{LSTM-CNN}
        \label{fig:results_detection_lstm}
    \end{subfigure}
    \caption{ Quantitative liquid detection results. The graphs indicate the precision and recall for each of the three networks. The colored lines indicate the variation in the number of slack pixels we allowed for prediction, i.e., how many pixels a positive classification could be away from a positive ground truth labeling and still be counted as correct.}
    \label{fig:results_detection}
\end{figure}

Figure \ref{fig:results} shows qualitative results for the three networks on the liquid detection task. 
The sequences used for this figure were randomly selected from the training set, and the frame with the most liquid visible was selected for display here\footnotemark[1].
It is clear from the figure that all three networks detect the liquid at least to some degree. 
The single frame CNN is less accurate at a pixel level, but it is still able to broadly detect the presence and general vicinity of liquid. 
As expected, the multi-frame CNN is much more precise than the single frame CNN.
Surprisingly, the LSTM CNN output appears much more accurate than even the multi-frame CNN.

Figure \ref{fig:results_detection} shows a quantitative comparison between the three networks. 
It plots the precision-recall curves for each of the networks when classifying each pixel as liquid.
We plot multiple lines for different amounts of ``slack,'' i.e., how many pixels a positive classification is allowed to be from a positive ground truth pixel and still count as correct.
We add this analysis due to the outputs shown in figure \ref{fig:results}; the networks clearly are able to detect the liquid, but are not necessarily pixel-for-pixel perfect. However, for the purposes of liquid manipulation, perfect pixel-wise accuracy is not necessary.

The quantitative results in figure \ref{fig:results_detection} confirm the qualitative outputs shown in figure \ref{fig:results}. 
As expected, the multi-frame CNN outperforms the single-frame. 
Surprisingly, the LSTM CNN performs much better than both by a significant margin, and even gets a significant boost from only a few slack pixels, indicating that even if the LSTM CNN is not always pixel-for-pixel accurate, it is often very close.
Taken together, these results strongly suggest that detecting transparent liquid must be done over a series of frames, rather than a single frame.

\subsection{Tracking Results}

\begin{figure}
    \centering
    \setlength{\unitlength}{1.0cm}
    \begin{subfigure}{6.0cm}
        \begin{picture}(6.0,5.0)
            \put(0.0,0.0){\includegraphics[width=6.0cm]{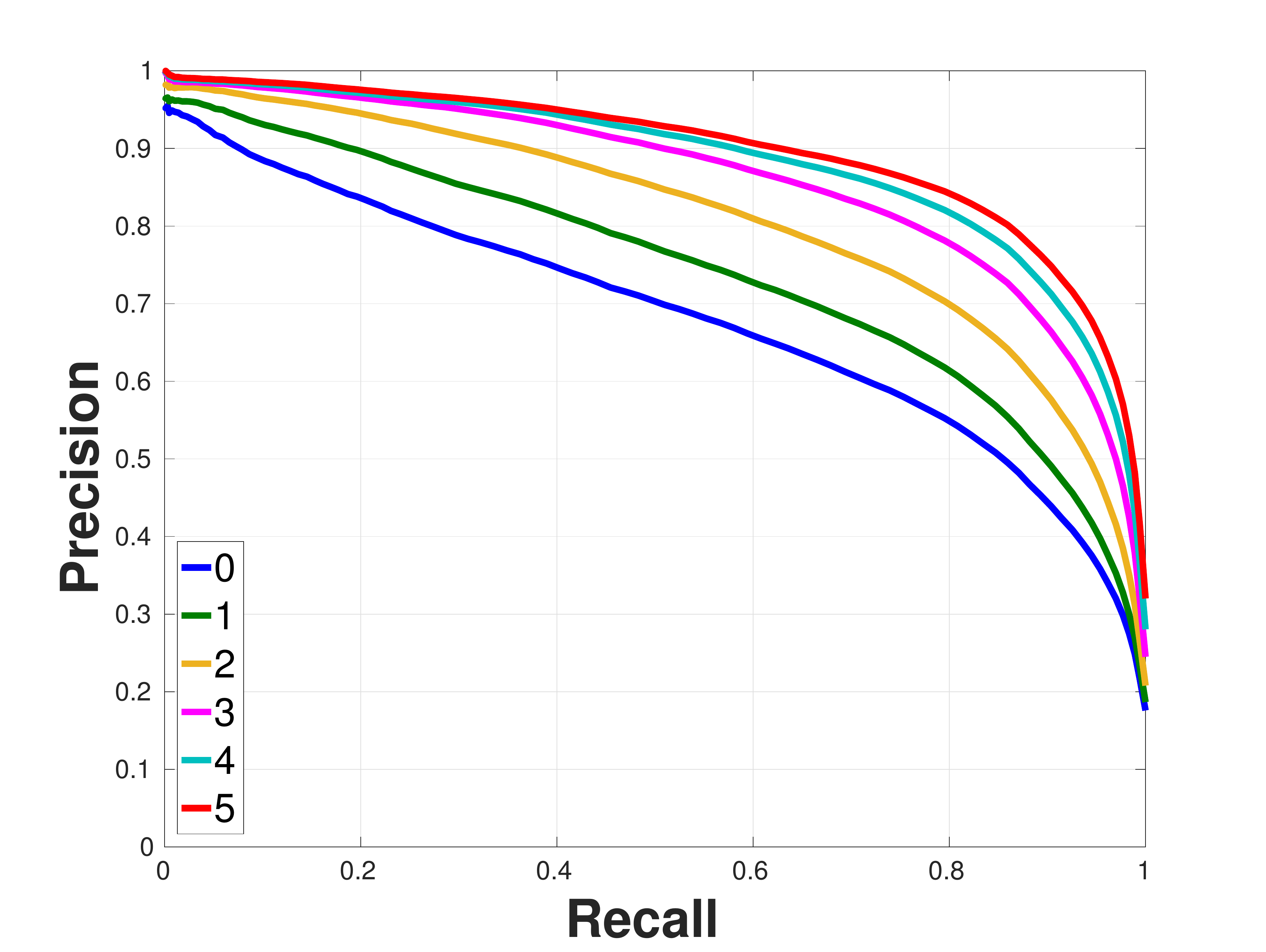}}
        \end{picture}
        \caption{CNN}
        \label{fig:results_tracking_cnn}
    \end{subfigure}
    
    \begin{subfigure}{6.0cm}
        \begin{picture}(6.0,5.0)
            \put(0.0,0.0){\includegraphics[width=6.0cm]{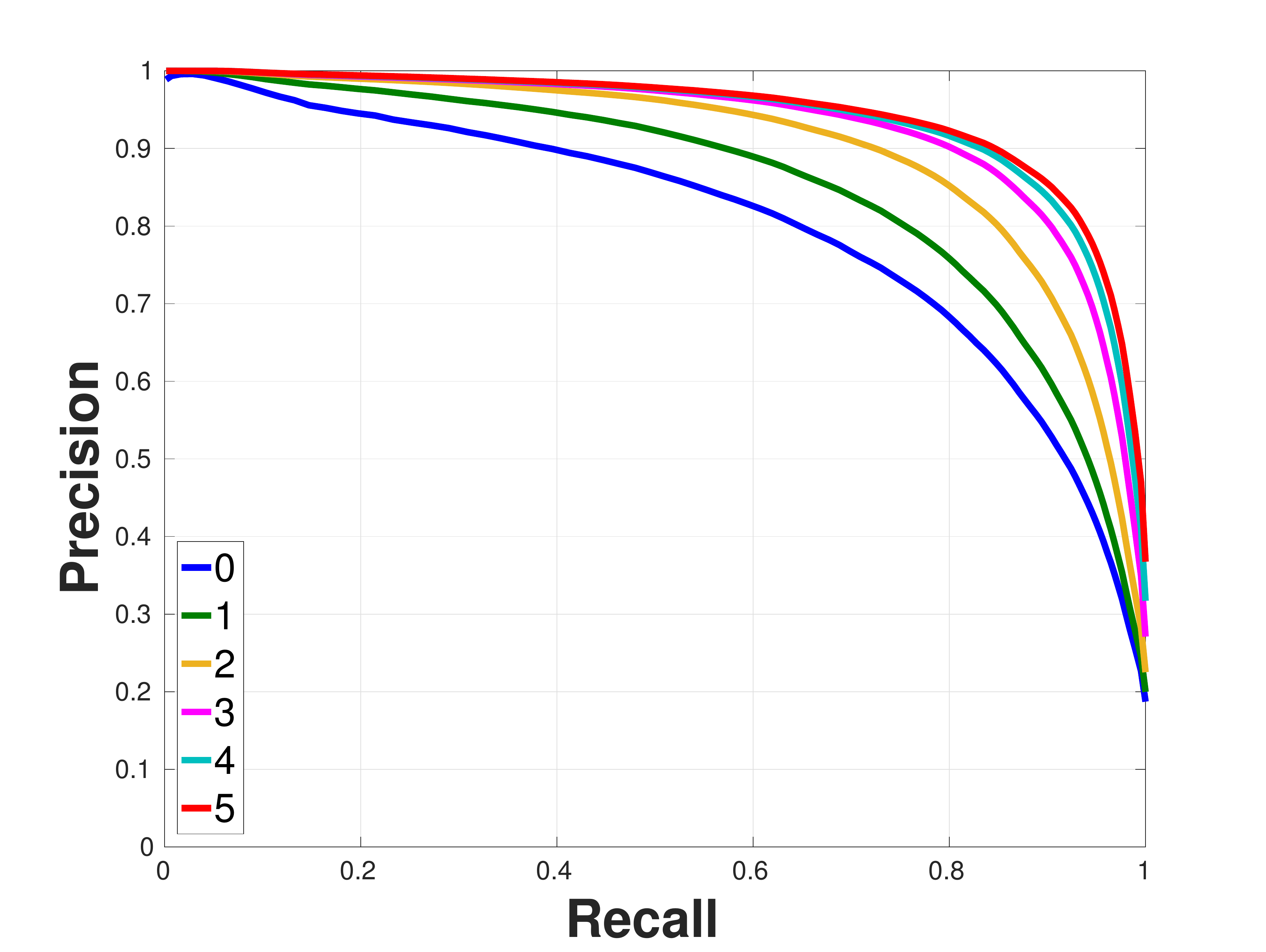}}
        \end{picture}
        \caption{MF-CNN}
        \label{fig:results_tracking_wcnn}
    \end{subfigure}
    
    \begin{subfigure}{6.0cm}
        \begin{picture}(6.0,5.0)
            \put(0.0,0.0){\includegraphics[width=6.0cm]{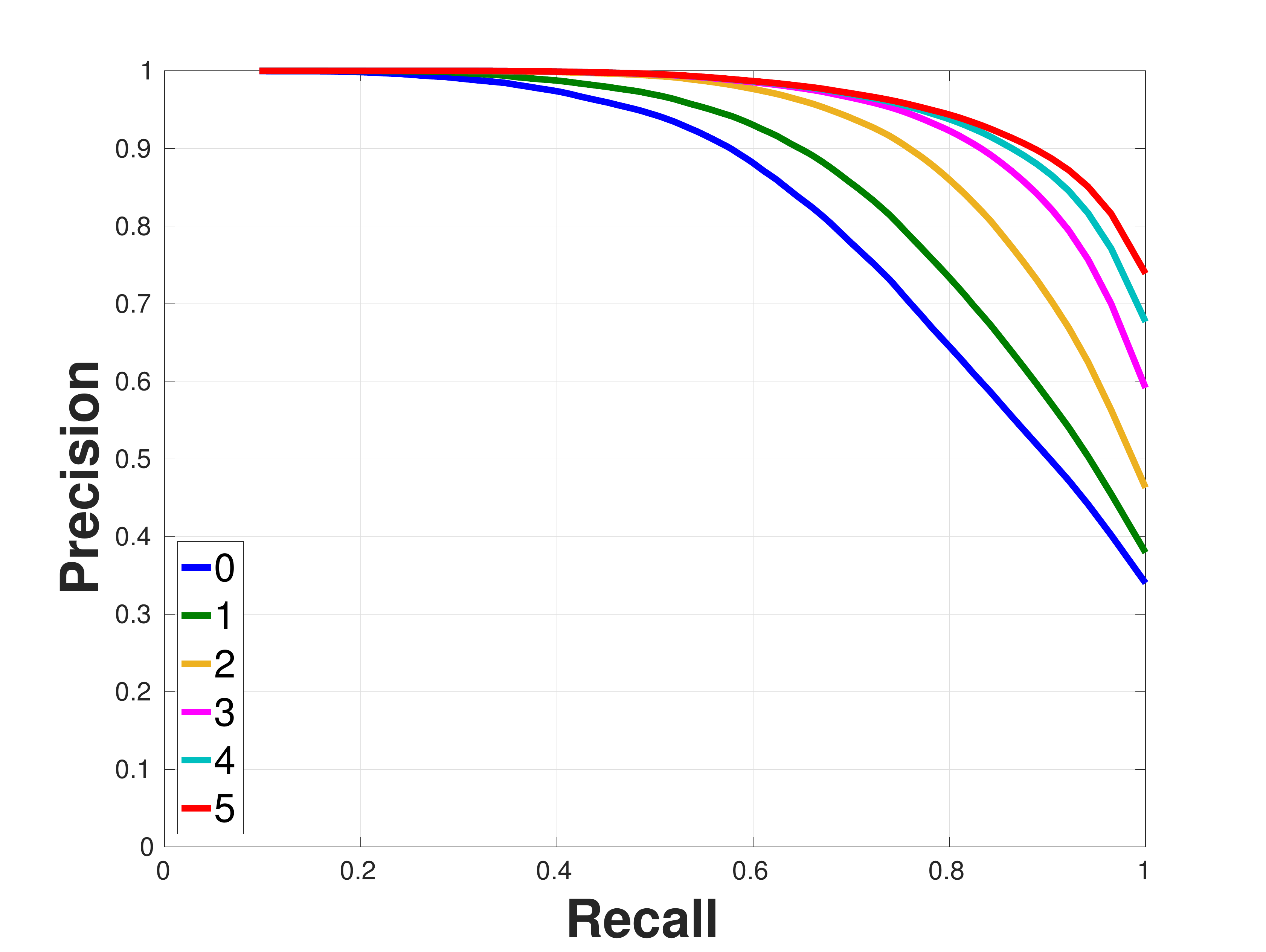}}
        \end{picture}
        \caption{LSTM-CNN}
        \label{fig:results_tracking_lstm}
    \end{subfigure}
    \caption{ Quantitative liquid tracking results. Similar to figure \ref{fig:results_detection}, the graphs indicate the precision and recall for each of the three networks and the colored lines indicate the variation in the number of slack pixels we allowed for prediction.}
    \label{fig:results_tracking}
\end{figure}

For tracking, we evaluated the performance of the networks on locating both visible and invisible liquid, given segmented input (i.e., each pixel classified as liquid, cup, bowl, or background). 
Because the viewpoint was fixed level with the bowl, the only visible liquid the network was given was liquid as it passed from cup to bowl. 
Figure \ref{fig:results_tracking} shows the performance of each of the three networks, and the accompanying video\footnotemark[1] shows the tracking results from the same 5 sequences shown in figure \ref{fig:results}.
Once again we plot the precision-recall curves for each network for different thresholds for the amount of ``slack'' given to each positive classification (i.e., the number of pixels a positive classification is allowed to be from a true positive pixel to count as correct).
As expected, the LSTM CNN has the best performance since it is the only network that has a memory, which is necessary to keep track of the occluded water in the cup and bowl.
Interestingly, the multi-frame CNN performs better than expected, given that it only sees approximately 1 second's worth of data and has no memory capability.
We suspect this is due to the network's ability to infer the likely location of the liquid based on the angle of the cup and the direction it's moving.
The video further reinforces this, as it is clear that in the sequence without liquid (the final sequence), the multi-frame CNN incorrectly infers the existence of water, whereas the LSTM CNN does not.

\section{Discussion \& Conclusion}
\label{sec:conclusion}

%Neural networks have the potential to be a key component for enabling robots to handle liquids using robust, closed-loop controllers.

%LSTMS work best, due to their ability to perform short term data integration (just like the MFCNN) AND to remember states, which is crucial for tracking the presence of water even when it's invisible.

The results in section \ref{sec:results} show that it is possible for deep learning to independently detect and track liquids in a scene. 
Unlike prior work on image segmentation, these results clearly show that single images are not sufficient to reliably perceive liquids. 
Intuitively, this makes sense, as a transparent liquid can only be perceived through its refractions, reflections, and specularities, which vary significantly from frame to frame, thus necessitating aggregating information over multiple frames. 
We also found that LSTM-based CNNs are best suited to not only aggregate this information, but also to track the liquid as it moves between containers. 
LSTMs work best, due to not only their ability to perform short term data integration (just like the MF-CNN), but also to remember states, which is crucial for tracking the presence of liquids even when they're invisible.

From the results shown in figure \ref{fig:results} and in the
video\footnotemark[1], it is clear that the LSTM CNN
can at least roughly detect and track liquids, although its pixel-wise accuracy
is not always 100\%, especially when the liquid is not visible.  Nevertheless,
unlike the task of image segmentation, our ultimate goal is not to perfectly
estimate the potential location of liquids, but to perceive and reason about the
liquid such that it is possible to manipulate it using raw sensory data.  For
this, a rough sense of where the liquid is in a scene and how it is moving might
suffice.  Neural networks, then, have the potential to be a key component for
enabling robots to handle liquids using robust, closed-loop controllers.

\footnotetext[1]{Video of the full sequences at \url{\youtubeurl}}

\section{Future Work}
\label{sec:future_work}

Further pursuing the problem of perceiving and reasoning about liquids, in future work we plan to combine the problems of detection and tracking into a single problem. 
Our goal is to not only perceive the liquid as it moves, but also to determine how much liquid is contained in the objects in the scene, and how much liquid is flowing. 
This is a necessary step before robots can apply control policies to liquid manipulation. 
The results here clearly show that the LSTM CNN is best suited for this task, and it is this type of network design we plan to investigate further for simultaneous detection and tracking of liquids from raw simulated imagery.

Another avenue for future work that we are currently pursuing is applying the techniques described here to data collected on a real robot. 
As stated in section \ref{sec:methodology}, it can be difficult to get the ground truth pixel labels for real data, which is why we chose to use a realistic liquid simulator in this paper. 
However, we are developing a method that uses a thermal infrared camera in combination with heated water to acquire ground truth labeling for data collected using a real robot.
The advantage of this method is that heated water appears identical to room temperature water on a standard color camera, but is easily distinguishable on a thermal camera.
This will allow us to label the ``hot'' pixels as liquid and all other pixels as not liquid.

Finally, we also plan to release not only our large dataset of labeled images, but also our code for generating this dataset. 
Other researchers will be able to apply their own algorithms to detecting and tracking liquid from raw sensory data. 
They will also be able to generate more data, and even vary how the data is generated (e.g., adding different types of cups, such as a glass cup). 
This will be the first dataset dedicated to perceiving and reasoning about liquids directly from raw sensory data generated via realistic simulation.

%This suggests that in future work it should be possible to combine the detection and tracking networks here with a planning or control algorithm in order to manipulate liquids using only raw sensory input.

\bibliographystyle{plainnat}
\bibliography{rss2016_ws}

\begin{thebibliography}{21}
\providecommand{\natexlab}[1]{#1}
\providecommand{\url}[1]{\texttt{#1}}
\expandafter\ifx\csname urlstyle\endcsname\relax
  \providecommand{\doi}[1]{doi: #1}\else
  \providecommand{\doi}{doi: \begingroup \urlstyle{rm}\Url}\fi

\bibitem[{Blender Online Community}(2016)]{blender2016}
{Blender Online Community}.
\newblock \emph{Blender - {A} 3D modelling and rendering package}.
\newblock Blender Foundation, Blender Institute, Amsterdam, 2016.
\newblock URL \url{http://www.blender.org}.

\bibitem[Cakmak and Thomaz(2012)]{cakmak2012}
Maya Cakmak and Andrea~L Thomaz.
\newblock Designing robot learners that ask good questions.
\newblock In \emph{ACM/IEEE International Conference on Human-Robot Interaction
  (HRI)}, pages 17--24, 2012.

\bibitem[Greff et~al.(2015)Greff, Srivastava, Koutn{\'\i}k, Steunebrink, and
  Schmidhuber]{greff2015}
Klaus Greff, Rupesh~Kumar Srivastava, Jan Koutn{\'\i}k, Bas~R Steunebrink, and
  J{\"u}rgen Schmidhuber.
\newblock Lstm: A search space odyssey.
\newblock \emph{arXiv preprint arXiv:1503.04069}, 2015.

\bibitem[Griffith et~al.(2012)Griffith, Sukhoy, Wegter, and
  Stoytchev]{griffith2012}
Shane Griffith, Vladimir Sukhoy, Todd Wegter, and Alexander Stoytchev.
\newblock Object categorization in the sink: Learning behavior--grounded object
  categories with water.
\newblock In \emph{Proceedings of the 2012 ICRA Workshop on Semantic
  Perception, Mapping and Exploration}. Citeseer, 2012.

\bibitem[Guo et~al.(2014)Guo, Singh, Lee, Lewis, and Wang]{guo2014}
Xiaoxiao Guo, Satinder Singh, Honglak Lee, Richard~L Lewis, and Xiaoshi Wang.
\newblock Deep learning for real-time atari game play using offline monte-carlo
  tree search planning.
\newblock In \emph{International Conference on Neural Information Processing
  Systems (NIPS)}, pages 3338--3346, 2014.

\bibitem[Havaei et~al.(2015)Havaei, Davy, Warde-Farley, Biard, Courville,
  Bengio, Pal, Jodoin, and Larochelle]{havaei2015}
Mohammad Havaei, Axel Davy, David Warde-Farley, Antoine Biard, Aaron Courville,
  Yoshua Bengio, Chris Pal, Pierre-Marc Jodoin, and Hugo Larochelle.
\newblock Brain tumor segmentation with deep neural networks.
\newblock \emph{arXiv preprint arXiv:1505.03540}, 2015.

\bibitem[Hochreiter and Schmidhuber(1997)]{hochreiter1997}
Sepp Hochreiter and J{\"u}rgen Schmidhuber.
\newblock Long short-term memory.
\newblock \emph{Neural computation}, 9\penalty0 (8):\penalty0 1735--1780, 1997.

\bibitem[Jia et~al.(2014)Jia, Shelhamer, Donahue, Karayev, Long, Girshick,
  Guadarrama, and Darrell]{jia2014}
Yangqing Jia, Evan Shelhamer, Jeff Donahue, Sergey Karayev, Jonathan Long, Ross
  Girshick, Sergio Guadarrama, and Trevor Darrell.
\newblock Caffe: Convolutional architecture for fast feature embedding.
\newblock \emph{arXiv preprint arXiv:1408.5093}, 2014.

\bibitem[K{\"o}rner et~al.(2006)K{\"o}rner, Pohl, R{\"u}de, Th{\"u}rey, and
  Zeiser]{korner2006}
Carolin K{\"o}rner, Thomas Pohl, Ulrich R{\"u}de, Nils Th{\"u}rey, and Thomas
  Zeiser.
\newblock Parallel lattice boltzmann methods for cfd applications.
\newblock In \emph{Numerical Solution of Partial Differential Equations on
  Parallel Computers}, pages 439--466. Springer, 2006.

\bibitem[Kunze and Beetz(2015)]{kunze2015}
Lars Kunze and Michael Beetz.
\newblock Envisioning the qualitative effects of robot manipulation actions
  using simulation-based projections.
\newblock \emph{Artificial Intelligence}, 2015.

\bibitem[Langsfeld et~al.(2014)Langsfeld, Kaipa, Gentili, Reggia, and
  Gupta]{langsfeld2014}
Joshua~D Langsfeld, Krishnanand~N Kaipa, Rodolphe~J Gentili, James~A Reggia,
  and Satyandra~K Gupta.
\newblock Incorporating failure-to-success transitions in imitation learning
  for a dynamic pouring task.
\newblock In \emph{IEEE International Conference on Intelligent Robots and
  Systems (IROS) Workshop on Compliant Manipulation}, 2014.

\bibitem[Levine et~al.(2015)Levine, Finn, Darrell, and Abbeel]{levine2015}
Sergey Levine, Chelsea Finn, Trevor Darrell, and Pieter Abbeel.
\newblock End-to-end training of deep visuomotor policies.
\newblock \emph{arXiv preprint arXiv:1504.00702}, 2015.

\bibitem[Long et~al.(2015)Long, Shelhamer, and Darrell]{long2015}
Jonathan Long, Evan Shelhamer, and Trevor Darrell.
\newblock Fully convolutional networks for semantic segmentation.
\newblock In \emph{IEEE International Conference on Computer Vision and Pattern
  Recognition (CVPR)}, pages 3431--3440, 2015.

\bibitem[Oh et~al.(2015)Oh, Guo, Lee, Lewis, and Singh]{junhyuk2015}
Junhyuk Oh, Xiaoxiao Guo, Honglak Lee, Richard~L Lewis, and Satinder Singh.
\newblock Action-conditional video prediction using deep networks in atari
  games.
\newblock In C.~Cortes, N.~D. Lawrence, D.~D. Lee, M.~Sugiyama, and R.~Garnett,
  editors, \emph{International Conference on Neural Information Processing
  Systems (NIPS)}, pages 2863--2871. 2015.

\bibitem[Okada et~al.(2006)Okada, Kojima, Sagawa, Ichino, Sato, and
  Inaba]{okada2006}
Kei Okada, Mitsuharu Kojima, Yuichi Sagawa, Toshiyuki Ichino, Kenji Sato, and
  Masayuki Inaba.
\newblock Vision based behavior verification system of humanoid robot for daily
  environment tasks.
\newblock In \emph{IEEE-RAS International Conference on Humanoid Robotics
  (Humanoids)}, pages 7--12, 2006.

\bibitem[Rankin and Matthies(2010)]{rankin2010}
Arturo Rankin and Larry Matthies.
\newblock Daytime water detection based on color variation.
\newblock In \emph{IEEE/RSJ International Conference on Intelligent Robots and
  Systems (IROS)}, pages 215--221, 2010.

\bibitem[Rankin et~al.(2011)Rankin, Matthies, and Bellutta]{rankin2011}
Arturo~L Rankin, Larry~H Matthies, and Paolo Bellutta.
\newblock Daytime water detection based on sky reflections.
\newblock In \emph{IEEE International Conference on Robotics and Automation
  (ICRA)}, pages 5329--5336, 2011.

\bibitem[Romera-Paredes and Torr(2015)]{romera2015}
Bernardino Romera-Paredes and Philip~HS Torr.
\newblock Recurrent instance segmentation.
\newblock \emph{arXiv preprint arXiv:1511.08250}, 2015.

\bibitem[Rozo et~al.(2013)Rozo, Jimenez, and Torras]{rozo2013}
Leonel Rozo, Pedro Jimenez, and Carme Torras.
\newblock Force-based robot learning of pouring skills using parametric hidden
  markov models.
\newblock In \emph{IEEE-RAS International Workshop on Robot Motion and Control
  (RoMoCo)}, pages 227--232, 2013.

\bibitem[Tamosiunaite et~al.(2011)Tamosiunaite, Nemec, Ude, and
  W{\"o}rg{\"o}tter]{tamosiunaite2011}
Minija Tamosiunaite, Bojan Nemec, Ale{\v{s}} Ude, and Florentin
  W{\"o}rg{\"o}tter.
\newblock Learning to pour with a robot arm combining goal and shape learning
  for dynamic movement primitives.
\newblock \emph{Robotics and Autonomous Systems}, 59\penalty0 (11):\penalty0
  910--922, 2011.

\bibitem[Yamaguchi and Atkeson(2015)]{yamaguchi2015}
Akihiko Yamaguchi and Christopher~G Atkeson.
\newblock Differential dynamic programming with temporally decomposed dynamics.
\newblock In \emph{IEEE-RAS International Conference on Humanoid Robotics
  (Humanoids)}, pages 696--703, 2015.

\end{thebibliography}

\end{document}